\UseRawInputEncoding

\documentclass[letterpaper, 10 pt, conference]{ieeeconf}  

\IEEEoverridecommandlockouts                              

\usepackage{graphicx}
\usepackage{tabularx}

\title{\LARGE \bf
Using Focus Group Interviews to Examine Biased Experiences in Human-Robot-Interaction
}

\author{Lukas Erle$^{1}$, Lara Timm$^{1}$, Carolin Stra{\ss}mann$^{1}$ and Sabrina C. Eimler$^{1}$,
\thanks{$^{1}$This research was funded by the German Federal Ministry of Education and Research as part of the RuhrBots project under the code RuhrBots-RA3P2.
All authors are with the Institute of Computer Science, Ruhr West University of Applied Sciences, Bottrop, Germany. 
 \{{lukas.erle, lara.timm, carolin.strassmann, sabrina.eimler}\}@hs-ruhrwest.de}%
}

\begin{document}

\maketitle
\thispagestyle{empty}
\pagestyle{empty}

\begin{abstract}
When deploying interactive agents like (social) robots in public spaces they need to be able to interact with a diverse audience, with members each having individual diversity characteristics and prior experiences with interactive systems. To cater for these various predispositions, it is important to examine what experiences citizens have made with interactive systems and how these experiences might create a bias towards such systems. To analyze these bias-inducing experiences, focus group interviews have been conducted to learn of citizens’ individual discrimination experiences, their attitudes towards and arguments for and against the deployment of social robots in public spaces. This extended abstract focuses especially on the method and measurement of diversity. 
\end{abstract}

\section{INTRODUCTION}

Social robots are a special category of interactive agents, which are usually designed with human-like appearances or abilities \cite{reeves_social_2020} and who interact with people in a social context, often assisting them in various ways \cite{gongora_alonso_social_2019, breazeal_social_2011}. One possible deployment field are public spaces, such as libraries, city administration offices 
\cite{ge_social_2018, casiddu_role_2019}. They can be used to instruct people and communicate important information, increase perceived safety and creativity in social interactions \cite{mintrom_robots_2022} and more generally act as embodied intermediaries \cite{schulz_web_2000} between citizens and public institutions. 

However, in these public spaces, social robots are often faced with a diverse audience as citizens exhibit unique combinations of diversity characteristics \cite{recchiuto_diversity-aware_2022}. Additionally, different citizens will have made different experiences with various technologies, some of them negative. These negative experiences can be multifaceted: Some citizens might have used certain devices or functions and simply encountered technological hurdles, user errors or incomplete or unreliable programs. At the same time, however, some citizens will already have encountered discriminations carried out by technological systems. These discriminations are often referred to as algorithmic bias, a term describing the existence of biases in algorithms and devices towards certain cultural, religious or other groups of people \cite{lambrecht_algorithmic_2016}. Those – whilst often not intended by a system’s developers \cite{hajian_algorithmic_2016} – can have a significant negative impact on affected people’s perception of and interaction with these systems. For example, people who have been subjected to algorithmic bias might evaluate the system as less fair and, as a result, do not trust its recommendations or even oppose using the system altogether \cite{kordzadeh_algorithmic_2022}. The decision to not use a system – especially when it is a public system like a social robot in a public space – then also forces citizens into a digital divide. This should be avoided at all costs, as it might lead to a reduced participation in public life \cite{norris_digital_2003, riggins_digital_2005}, as well as negatively impact social proximity of citizens \cite{rossler_digital_2017}. 

The goal of this research is to examine what experiences citizens have made with different technologies and how instances of algorithmic bias might lead to biased interactions with social robots, as well as how they behave in biased interactions and what strategies might be used to cope with such bias. To this end we are conducting multiple focus group interviews with citizens of the Ruhr area in Germany, thereby gathering individual experiences, hurdles, motivators, and coping strategies for regular and biased interactions with social robots.

\section{DIVERSITY IN HRI RESEARCH}

Diversity has found different definitions across extant literature. Generally, diversity is split into two subcategories \cite{de_graaf_inclusive_2022}, the first of which being activity-based diversity. This subcategory differentiates people purely based on their occupation. The second subcategory, relational diversity, describes people’s ethnic origin, religious affiliations and other – often unchangeable – aspects of their identity. In line with \cite{de_graaf_inclusive_2022} and \cite{mouret_encouraging_2012}, we argue that personal identity as a whole consists of both these subcategories and therefore needs to be treated as a complex construct that requires further definition and examination \cite{hutchison_towards_2012}.  

While extant literature agrees that diversity is a relevant factor in the design and conceptualization of virtual agents \cite{baranauskas_characterizing_2007, simpson_measurement_1949, poulain_quantifying_2018}, few research has been carried out on how to quantify and measure users’ diversity in HRI research. Different approaches have been developed over the years: For example, modelling techniques such as the \textit{repertory grid technique} or \textit{multi-dimensional scaling} can be used to analyze users’ perceptions about a system \cite{mccrae_introduction_1992}. However, these models share a common shortcoming: They are largely quantitative ways of describing and defining diverse user groups, often in the context of a concrete system. As a result, the qualitative nature of diversity characteristics is ignored by those models.

A more fitting concept aimed at describing humans’ diversity characteristics is a representation of those characteristics in the form of a diversity wheel \cite{gardenswartz_diverse_2003}. 
Its four layers of diversity are \textit{personality, internal dimensions, external dimensions,} and \textit{organizational dimensions}. In order of mention, these dimensions become more and more flexible and changeable: Organizational dimensions like \textit{work content} or \textit{department} can more easily be changed by an individual than their external dimensions, such as their educational background or appearance. Vitally, the internal dimensions also describe diversity characteristics that are almost immutable, for example (biological) gender, sexual orientation, and age. The diversity wheel thereby summarizes both activity-based diversity characteristics (organizational dimensions) and relational diversity characteristics (external and internal dimensions), while also adding an individual’s personality at the core. Initially developed for a business context, the wheel has been used in recent research on diversity in companies and organizations \cite{sihn-weber_diversity_2021, kaudela-baum_diversity-kompetent_2022}. Since the wheel has been used only occasionally in HRI research, we intend to focus it more strongly. This research therefore follows a novel approach by attempting to use its dimensions to describe the diversity characteristics of users interacting with social robots. 

\section{METHOD}

To carry out our research agenda, we decided to conduct focus group interviews with different citizens from the Ruhr area in Germany. Focus groups have been proven to contribute to the understanding of multiple opinions and experiences regarding a certain topic \cite{brits_application_2007, dilshad_focus_2013} while also allowing unique new perspectives through discussions between participants \cite{jackson_focus_1998}. They therefore are a suitable method for examining citizens’ experiences with and attitudes towards modern technologies. Because the specific functions of social robots in a public space might be subject to change depending on the situation and space they are being deployed in, we decided to not only examine the participants’ experiences with and attitudes towards (social) robots, but modern technologies and functions more generally. Specifically, we aggregated a list of popular devices and functions. For devices, we asked examined participants’ experience with \textit{laptops/computers, smartphones, tablets, smartwatches, VR/AR glasses, chatbots, phone bots, voice assistants, robots, touch terminals, digital cameras, TVs,} and \textit{e-readers}. For functions, we examined \textit{facial recognition, voice recognition, voice and video calls, fingerprint recognition,} and \textit{head-tracking}. 

\subsection{Focus Group Setup}
The focus group interviews were planned in groups of eight participants each (which was not possible in every focus group due to last minute cancellations). The first cohort of participants consisted of university students and was recruited through the university’s e-learning platform and divided into different time slots. Each focus group interview was scheduled for two and a half hours, their audio was recorded, and they followed a semi-structured interview guide. This method was chosen to ensure a comparable proceeding of each focus group interview whilst still allowing us to dive deeper into some of the participants’ experiences if needed. For this purpose, semi-structured interview guides have proven to be a suitable method in various disciplines \cite{wilson_semi-structured_2014, kallio_systematic_2016, horton_qualitative_2004}. Participants were provided with a printed booklet, including blank sheets with the different questions asked as part of the interview guide and a demographic questionnaire. To be able to connect the participants’ written answers and demographic data to their verbal expressions without revealing their identity, we asked them to choose one of 12 different superhero identities. Participants then wrote their superhero name on each page of the booklet and were only addressed with their superhero name throughout the interview. Participants were first assigned their superhero identities and were then briefed about the contents and procedure of the interview and signed a declaration of consent, allowing us to record the interview and using these recordings for further analysis. 

\subsection{Measuring Prior Experiences and Personality}
To gauge participants’ individual attitudes towards and experiences with technologies – and specifically social robots – we decided to measure these both quantitatively and qualitatively. For the quantitative part, participants were asked to fill out various questionnaires: Participants’ attitudes towards robots were measured using the \textit{General Attitudes towards Robots Scale (GAToRS)} \cite{koverola_general_2022}. For application with German participants, we translated the items form English to German and had multiple researchers translate those items back to English to ensure they were a suitable translation. For measuring participants’ general readiness to try out and interact with technology, we used the German short scale for \textit{technology commitment} \cite{neyer_kurzskala_2016}. Finally, we measured participants’ personality characteristics using the \textit{Big Five Inventory (BFI-10) scale} \cite{rammstedt_big_2014}. 

\subsection{Application of the Diversity Wheel}
As a final aspect to the participants’ demographic data, we wanted to examine their diversity characteristics. As hinted at in the previous section, we chose the diversity wheel to aid us in this endeavor. To ensure that the wheel and its dimensions are a good fit for the examination of social robots in public spaces, we chose a German translation of the original diversity wheel \cite{sihn-weber_diversity_2021, kaudela-baum_diversity-kompetent_2022} and critically assessed which dimensions would be relevant for the interaction with social robots in public spaces. We decided to remove the aspects \textit{Personal Habits, Recreational Habits, Work Experience, Appearance, Geographic Location, Functional Level, Division/Department, Seniority, Work Location, Union Affiliation,} and \textit{Management Status}. The reason behind the removal of those exact aspects from the wheel is the assumption that a social robot would not know these aspects about a citizen in a regular interaction, and some aspects might not be relevant to the interaction at all. For example, a citizen’s personal habits might be different or non-existent when being in a public space and interacting with a social robot there. Similarly, it would be very unlikely that a social robot would know of a citizen’s union affiliation, considering an employee is not even obligated to share their union affiliation with their employer. Each remaining aspect of the diversity wheel was then included as part of the demographic data questionnaire. We followed various standards and guidelines for capturing these aspects \cite{statistisches_bundesamt_demografische_2016, abdul-hussain_dimensionen_2013, technische_universitat_dresden_fragebogen_2018} and to keep anonymity whilst still creating a reliable and quantifiable picture of the participants’ diversity. At the end of this, participants were provided with the diversity wheel and asked to mark the three aspects that they deemed most important to their own identity. These pages were collected, and the aspects were anonymously transferred to a larger print of that same wheel, which was then hanged up for all participants to see. This was done to establish a common ground regarding the diversity characteristics represented in the group. Establishing a common ground has proven to be an important step to ensure questions can be understood correctly and answered precisely \cite{gibbs_common_1988}.

\subsection{Interview Guide}
 After filling out the aforementioned demographics questionnaire, participants were asked to state how often they used the devices and technologies under consideration (\textit{never to more than once a day}) and how they would rate them (\textit{positively, neutrally, negatively}). For this, large tables were prepared that allowed participants to state their usage frequency and evaluation by placing a dot in the corresponding field. 
This position in the interview guide represents the end of the quantitative part of the focus group interviews. To examine participants’ evaluation of the devices and functions in more detail, they were then asked to write down whether they have had any negative key experiences that made them dislike a device. When participants finished writing down their thoughts, they were asked to – voluntarily – share some of their experiences with the group. This approach was chosen to ensure that any intimidating effects of the other group members, the interview situation, or the moderators would not lead to any apprehensions in replies, whilst still allowing a public discussion of some negative experiences. Furthermore, this question was specifically phrased to prompt participants to share any experiences with algorithmic bias or discrimination by devices or functions. Next, this procedure was repeated, only this time asking participants to write down and share any positive experiences that were essential for their evaluation of devices or functions.
Afterwards, participants were divided into groups of two and – in two rounds – each received a scenario of algorithmic bias which had been printed out. There were eight scenarios in total, with five scenarios following real-world cases of algorithmic bias \cite{wulf_automatisierte_2022, orwat_diskriminierungsrisiken_2019}. The remaining three scenarios were constructed from deliberations made during an ethics workshop dealing with ethical implications of a deployment of social robots in a public space. The groups were given 15 minutes to read and discuss the scenario, with the goal to determine how the error transpired, whose fault it was and which technical solutions or coping strategies could be adopted to rectify the issue. Again, participants were first required to write down their answers and thoughts, with a public discussion following once the 15 minutes had run out. Each group then summarized their scenario to the other groups and shared their thoughts on the attribution of guilt and reason behind the issue. Other groups also had the chance to offer their thoughts on the summarized scenario. The whole process was repeated with a second set of scenarios until each group had worked on two scenarios. In this part of the interview, we wanted to examine whether the participants would attribute the issue to user error, to a malfunction, or recognize that the issue happened because of the users’ diversity characteristics. Furthermore, we were interested in how participants would behave in place of the users and what technical solutions they would envision to avoid these problems from happening in the future. Thereby, participants have also been made aware of and primed for possible algorithmic biases. For the final part of the interview, the moderators presented a scenario to the entirety of the group: 

\textit{You want to borrow a book from the public library in your city. You no longer interact with humans in the library but are instead accompanied by a robot during your visit. This robot can read stories to you, navigate you through the library and serve as an information terminal.  It could also perform the functions discussed earlier. It also understands human language and can respond both verbally and via a tablet.}

With this scenario set, participants were urged to fantasize about the best-case (utopian) and worst-case (dystopian) interactions with this social robot and, again, first write down their thoughts and share them with the rest of the group afterwards. 
At the end of the focus group interview, the filled-out booklets were collected, and – along with the tables – scanned and digitalized. The recordings of the interviews archived and transcribed for further analysis.

\section{CONCLUSION}

This extended abstract has introduced a novel approach of measuring diversity characteristics in HRI research in the application of the diversity wheel developed by \cite{gardenswartz_diverse_2003}. Furthermore, this approach has been applied to specific research on social robots in the form of focus group interviews. These interviews aimed at compiling citizens' experiences with modern technologies, specifically experiences with algorithmic bias. Further steps will include carrying out more focus groups interviews following with more diverse groups of citizens, as well as analysing the data gathered during these focus group interviews. At the end of these steps, we aim to reliably predict citizens' reaction to a deployment of social robots in public spaces and ensure that these social robots are suitable for interaction with diverse audiences.

\bibliographystyle{ieeetr}
\bibliography{sample-base}

\end{document}